\documentclass[11pt,a4paper]{article}
\usepackage[hyperref]{eacl2021}
\usepackage{times}
\usepackage{latexsym}

\usepackage{float}

\usepackage{amssymb}
\usepackage{booktabs}
\usepackage{xspace}
\newcommand{\acro}{MLQE-PE\xspace}
\usepackage{graphicx}
\usepackage{subcaption}
\usepackage{CJKutf8}
\usepackage{multirow}

\usepackage{microtype}

\aclfinalcopy 

\title{\acro: A Multilingual Quality Estimation and Post-Editing Dataset}

\author{\\
 Marina Fomicheva,\textsuperscript{1*} Shuo Sun,\textsuperscript{2*} Erick Fonseca,\textsuperscript{3} 
 Chrysoula Zerva,\textsuperscript{3} 
 Fr\'{e}d\'{e}ric Blain,\textsuperscript{1,4}\\
 Vishrav Chaudhary,\textsuperscript{5} 
 Francisco Guzm\'an,\textsuperscript{5}
 Nina Lopatina,\textsuperscript{6}
 Lucia Specia,\textsuperscript{1,7}Andr\'{e} F.~T. Martins\textsuperscript{3,8} \vspace{0.2cm}
\\
 \textsuperscript{1}University of Sheffield, 
 \textsuperscript{2}Johns Hopkins University, \\
 \textsuperscript{3}Instituto de Telecomunica\c{c}\~{o}es, 
 \textsuperscript{4}University of Wolverhampton, \\
 \textsuperscript{5}Facebook AI, 
 \textsuperscript{6}IQT Labs,
 \textsuperscript{7}Imperial College London,
  \textsuperscript{8}Unbabel\\
  \texttt{\{m.fomicheva,l.specia\}@sheffield.ac.uk}, \texttt{ssun32@jhu.edu}, \texttt{nlopatina@iqt.org}\\ 
  \texttt{erick.fonseca@lx.it.pt}, \texttt{f.blain@wlv.ac.uk}, \texttt{\{fguzman,vishrav\}@fb.com}, \\
  \texttt{chrysoula.zerva@tecnico.ulisboa.pt}, \texttt{andre.martins@unbabel.com}
}

\newif\ifdraft 
\draftfalse
\newcommand{\edit}[1]{\ifdraft{\textcolor{blue}{#1}}\else{#1}\fi}

\newcommand{\tagok}{\textsc{ok}}
\newcommand{\tagbad}{\textsc{bad}}

\begin{document}

\maketitle
\begin{abstract}
We present \acro, a new dataset for Machine Translation (MT) Quality Estimation (QE) and Automatic Post-Editing (APE). The dataset contains \edit{eleven} language pairs, with human labels for \edit{up to 10,000} translations per language pair in the following formats: sentence-level direct assessments and post-editing effort, and word-level good/bad labels. It also contains the post-edited sentences, 
as well as titles of the articles where the sentences were extracted from, and the neural MT models used to translate the text. 
\end{abstract}

\renewcommand{\thefootnote}{$^{*}$}
\footnotetext[1]{Equal contribution.}
\renewcommand\thefootnote{\arabic{footnote}}

\section{Introduction}



Translation quality estimation (QE) is the task of evaluating a translation system's quality without access to reference translations \citep{Blatz2004, Specia2018}. This task has numerous applications: deciding if a sentence or document that has been automatically translated is ready to be sent to the final user or if it needs to be post-edited by a human, flagging passages with potentially critical mistakes, using it as a metric for translation quality when a human reference is not available, or in the context of computer-aided translation interfaces, highlighting text that needs human revision and estimating the human effort. 

Due to its high relevance, QE has been the subject of evaluation campaigns in the Conference for Machine Translation (WMT) since 2014  \citep{bojar-EtAl:2014:W14-33,specia2018findings,fonseca-etal-2019-findings,specia-etal-2020-findings-wmt}, where datasets in various language pairs have been created containing source sentences, their automatic translations, and human post-edited text. However, the currently existing data has several shortcomings. First, the MT system used to produce the translations is not publicly available, 
which makes it impossible to develop the so-called glass-box approaches to QE and exploit model confidence (or conversely, uncertainty) of the MT system or look into its internal states. Second, the quality assessments have been either produced 
based on the difference between the MT output and the post-edited text (e.g., through the human translation error rate metric, HTER, or by marking individual words with {\sc ok} or {\sc bad} labels), or by direct human assessments, but not both---which raises the question of how much these two quality assessments correlate. Third, most datasets have focused exclusively on high-resource language pairs, where it is often the case that many sentences are correctly translated; however, medium and low-resource settings are the ones where QE would be particularly useful, since it is where MT currently presents serious challenges. Finally, most of these datasets focus on a specific domain, such as IT or life sciences, where translations are generated by a domain-specific MT model, which also tends to result in most sentences being translated with high-quality. 

To overcome the limitations stated above, we introduce \acro, the first multilingual quality estimation and post-editing dataset that combines the following features:
\begin{itemize}
    \item It includes access to the state-of-the-art neural MT (NMT) models built with an open-source toolkit ({\tt fairseq}, \citet{ott2019fairseq}), 
    that were used to produce the translations in the dataset. This opens the door to uncertainty-based and glass-box approaches to QE.
    \item It combines both direct assessments of MT quality and post-edits. This allows combining two sorts of quality assessments: how good a translation is and how much effort is necessary to correct it. Moreover, the post-edited sentences can be used for training and evaluating automatic post-editing systems, another important task considered in WMT campaigns \cite{chatterjee2019findings}.
    \item It contains the titles of the Wikipedia articles where the original sentences were extracted from, thus allowing to take document-level context into account when predicting sentence-level or word-level MT quality.
    \item It includes \edit{11} language pairs, mixing high-resource language pairs (English-German -- En-De and English-Chinese -- En-Zh, and Russian-English -- Ru-En), medium-resource (Romanian-English -- Ro-En, and Estonian-English -- Et-En) and low-resource ones (Nepali-English -- Ne-En, Sinhala-English -- Si-En, \edit{Pashto-English -- Ps-En, Khmer-English -- Km-En, English-Japanese -- En-Ja, and English-Czech -- En-Cs}).
\end{itemize}

\edit{This dataset was created with contributions from different institutions: Facebook, University of Sheffield and Imperial College selected the Wikipedia articles and sentences, built the NMT models, prepared and outsourced data for DA annotation in \edit{10} language pairs (En-De, En-Zh, Ro-En, Et-En, Ne-En, Si-En, \edit{Ps-En, Km-En, En-Ja, En-Cs}). IQT Labs led the same efforts for collecting and DA-annotating the Ru-En data. Facebook, University of Sheffield and Imperial College also outsourced data for all language pairs except En-De and En-Zh for post-editing, and created reference translations for Et-En. Unbabel and Instituto de Telecomunica\c{c}\~oes outsourced the post-editing of En-De and En-Zh sentences. }
The current version of \acro is publicly available from  \url{https://github.com/sheffieldnlp/mlqe-pe}. 

\section{Data Collection and Statistics}

We briefly describe the data collection and preparation process. Table~\ref{tab:data_statistics} presents some statistics about the \acro dataset. \edit{As shown in Table~\ref{tab:data_statistics}, we collected 10K sentences split into train, dev and two test partitions (test20 and test21) for nine language pairs. In addition, we collected 2K sentences for 4 language pairs, which are meant to be used for testing QE in a zero-shot setting where no training or development data is provided.\footnote{1K of these sentences will be kept as a blind test set and released later.}} 

\paragraph{Data collection.} 
\begin{table*}[t]
    \centering
    \begin{tabular}{lcccc}
        \toprule
        Languages & Sentences & Tokens & DA & PE \\
        \midrule
         En-De & 7,000/1,000/1,000 & 114,980 / 16,519 / 16,371 / 16,545 & \checkmark & \checkmark \\
         En-Zh & 7,000/1,000/1,000 & 115,585 / 16,307 / 16,765 / 16,637 & \checkmark & \checkmark \\
         Ru-En & 7,000/1,000/1,000 & 82,229 / 11,992 / 11,760 / 11,650 & \checkmark & \checkmark \\
         Ro-En & 7,000/1,000/1,000 & 120,198 / 17,268 / 17,001 / 17,359 & \checkmark & \checkmark \\
         Et-En & 7,000/1,000/1,000 & 98,080 / 14,423 / 14,358 / 14,044 & \checkmark & \checkmark \\
         Ne-En & 7,000/1,000/1,000 & 104,934 / 15,144 / 14,770 / 15,017 & \checkmark & \checkmark \\
         Si-En & 7,000/1,000/1,000 & 109,515 / 15,708 / 15,821 / 15,709 & \checkmark & \checkmark \\
         Ps-En & 1,000 & 27,045 & \checkmark & \checkmark \\
         Km-En & 1,000 & 21,981 & \checkmark & \checkmark \\
         En-Ja & 1,000 & 20,626 & \checkmark & \checkmark \\
         En-Cs & 1,000 & 20,394 & \checkmark & \checkmark \\
        \bottomrule
    \end{tabular}
    \caption{Statistics of the \acro dataset. The numbers of sentences and tokens are shown for train, development and two test partitions (test20 and test21), respectively for En-De, En-Zh, Ru-En, Ro-En, Et-En, Ne-En and Si-En, and for the test partition for Ps-En, Km-En, En-Ja and En-Cs. The number of tokens refers to the source side.}
    \label{tab:data_statistics}
\end{table*}

\begin{table}[t]
    \centering
    \begin{tabular}{lcc}
        \toprule
        & Average DA $\uparrow$ & Average HTER $\downarrow$ \\
        \midrule
        En-De & 82.61 & 0.18 \\
        Ro-En & 69.18 & 0.24 \\
        En-Ja & 67.96 & 0.36 \\
        En-Cs & 66.94 & 0.26 \\
        En-Zh & 62.86 & 0.23 \\
        Et-En & 60.09 & 0.29 \\
        Ps-En & 53.53 & 0.53 \\
        Si-En & 51.42 & 0.59 \\
        Km-En & 46.58 & 0.65 \\
        Ne-En & 36.51 & 0.66 \\
        \bottomrule
    \end{tabular}
    \caption{Average MT quality in terms of DA scores (higher is better) and HTER scores (lower is better) on the test21 partition of the dataset.}
    \label{tab:quality_stats}
\end{table}


For the most part, the dataset is derived from Wikipedia articles (with exception of Russian-English, described below). The source sentences were collected from Wikipedia articles following the sampling process outlined in FLORES \cite{guzman-etal-2019-flores}. First, we sampled documents from Wikipedia for English, Estonian, Romanian, Sinhala, Nepali, \edit{Khmer and Pashto}. Second, we selected the top 100 documents containing the largest number of sentences that are: (i) in the intended source language according to a language-id classifier\footnote{\url{https://fasttext.cc}} and (ii) have the length between 50 and 150 characters. In addition, we filtered out sentences that have been released as part of recent Wikipedia parallel corpora \cite{schwenk2019wikimatrix}, ensuring that our dataset is not part of parallel data commonly used for NMT training.

For every language, we randomly selected the required number of sentences from the sampled documents and then translated them using SOTA NMT models (see below). 
For German and Chinese, we followed an additional procedure in order to ensure sufficient representation of high- and low-quality translations for these high-resource language pairs. We selected the sentences with minimal lexical overlap with respect to the NMT training data. Specifically, we extracted content words for each sentence in the data used for training the NMT models and in the Wikipedia data. We computed perplexity scores for the Wikipedia sentences given the NMT training data. Finally, we sampled 20K from available Wikipedia sentences weighted by the perplexity scores.

In addition, we collected human reference translations for a 1K subset of Estonian-English dev/test data. Two reference translations were generated independently by two professional translators. This part of the dataset allows for comparing reference-free MT evaluation with reference-based approaches (see \citet{fomicheva2020multi} for details).


\begin{figure*}[ht]
\centering
    \includegraphics[width=.7\textwidth]{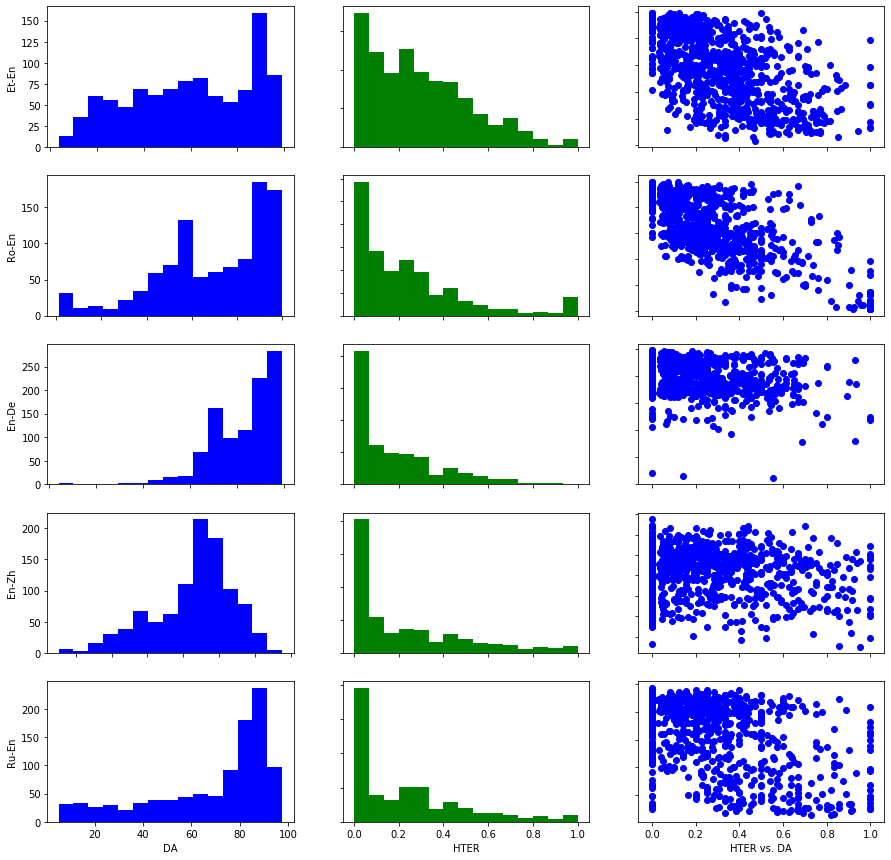}
\caption{Distribution of direct assessment scores (DA), HTER scores and their scatter plots for the test21 partition of the dataset, for Et-En, Ro-En, En-De, En-Zh and Ru-En language pairs.}
\label{fig:histograms_high_resource_test21}
\end{figure*}

\begin{figure*}[ht]
\centering
    \includegraphics[width=.7\textwidth]{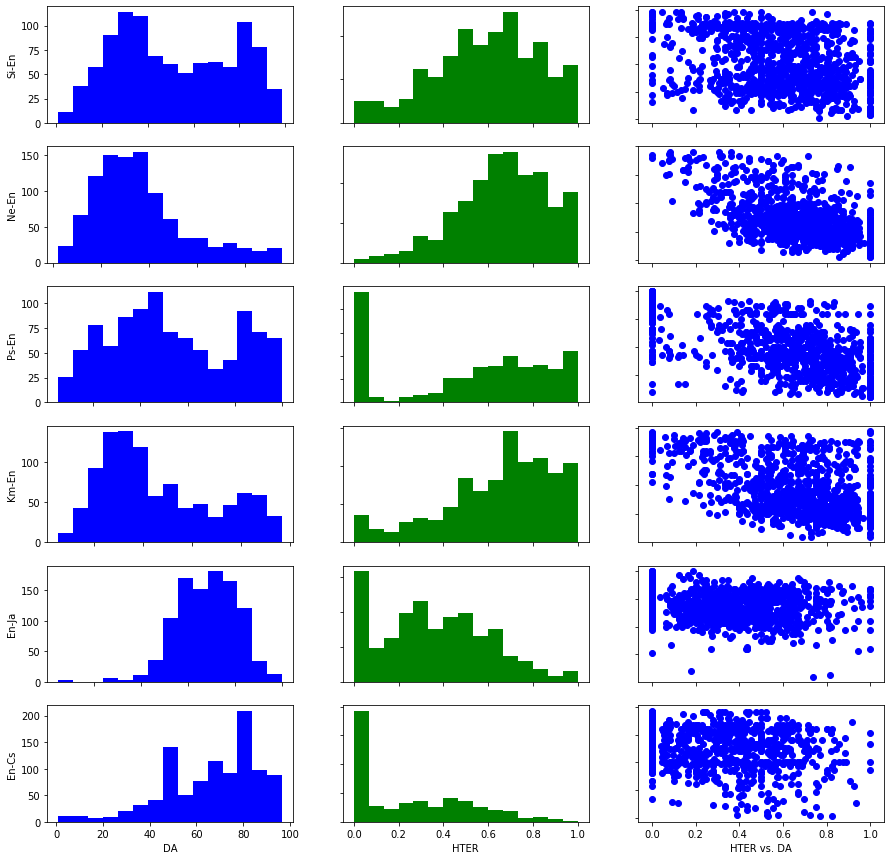}
\caption{Distribution of direct assessments scores (DA), HTER scores and their scatter plots for the test21 partition of the dataset, for Si-En, Ne-En, Ps-En, Km-En, En-Ja and En-Cs language pairs.}
\label{fig:histograms_low_resource_test21}
\end{figure*}

The Russian-English data collection followed a slightly different set up collected by collaborators from IQT Labs.\footnote{We note that Facebook was not involved in the collection of the Russian-English data.} The original sentences were collected from multiple sources in order to gather a varied sample of data in different domains that are still challenging for current NMT systems. Data sources include: Russian proverbs and Reddit data from various subreddits, particularly those focused on topics of politics and religion. We included Reddit data since colloquial text is a challenge for MT. We included Russian proverbs from WikiQuotes to test MT on short sentences with unconventional grammar. We used the Reddit API and queried the most recent 1000 posts at the time, and the most recent 1000 comments in each of the selected subreddits. We automatically split the posts into sentences and then reviewed these manually. Markdown was removed and HTML unencoded. We removed sentences shorter than 15 characters or longer than 500 characters. We also removed sentences that did not have a source link. Table \ref{tab:ruen-data-sources} shows the number of segments corresponding to each data source and the corresponding average direct assessment score.

\begin{table*}[t]
    \centering
    \begin{tabular}{lcc}
        \toprule
        & Count & DA \\
        \midrule
         www.reddit.com/r/antireligious & 2,155 & 75.6 \\ 
         www.reddit.com/r/PikabuPolitics & 1,753 & 77.7 \\ 
         www.reddit.com/r/rupolitika & 1,422 & 80.1 \\ 
         www.reddit.com/r/ru & 2,171 & 74.0 \\ 
         wikiquote.org/wiki & 2,499 & 41.1 \\ 
        \bottomrule
    \end{tabular}
    \caption{Number of sentences and average absolute direct assessment (DA) score for each data source in the Ru-En dataset}
    \label{tab:ruen-data-sources}
\end{table*}

\begin{table}[t]
    \centering
    \begin{tabular}{lcc}
        \toprule
        & Pearson & Spearman\\
        \midrule
		En-De & -0.42 & -0.48\\
		Ro-En & -0.76 & -0.71\\
		En-Ja & -0.14 & -0.11\\
		En-Cs & -0.41 & -0.46\\
		En-Zh & -0.21 & -0.16\\
		Et-En & -0.61 & -0.63\\
		Ps-En & -0.71 & -0.67\\
		Si-En & -0.29 & -0.28\\
		Km-En & -0.49 & -0.43\\
		Ne-En & -0.54 & -0.49\\
        \bottomrule
    \end{tabular}
    \caption{Pearson and Spearman correlation between DA and HTER scores for the test21 partition of the dataset.}
    \label{tab:da_hter_correlation}
\end{table}





\paragraph{NMT models}

Transformer-based \cite{vaswani2017attention} NMT models were trained for all languages using the {\tt fairseq} toolkit.\footnote{\url{https://github.com/pytorch/fairseq}} For \textbf{Et-En}, \textbf{Ro-En}, \textbf{En-De} and \textbf{En-Zh} we trained the MT models based on the standard Transformer architecture following the implementation details described in \citet{ott2018scaling}. We used publicly available MT datasets such as Paracrawl \cite{espla-etal-2019-paracrawl} and Europarl \cite{koehn2005europarl}. For \textbf{Ru-En}, translations were produced with the already existing Transformer-based NMT model described in \citet{fair-wmt2019}.\footnote{\url{https://github.com/pytorch/fairseq/tree/master/examples/wmt19}} \textbf{Si-En} and \textbf{Ne-En} MT systems were trained based on Big-Transformer architecture as defined in \citet{vaswani2017attention}. For these low-resource language pairs, the models were trained following the FLORES semi-supervised setting \cite{guzman-etal-2019-flores},\footnote{\url{https://github.com/facebookresearch/flores/blob/master/reproduce.sh}} which involves two iterations of backtranslation using the source and the target monolingual data. \edit{For \textbf{Ps-En}, \textbf{Km-En}, \textbf{En-Cs} and \textbf{En-Ja} we use multilingual MT models described in \citet{Tang2020MultilingualTW}.\footnote{Instructions for training the models and generating the translations can be found at \url{https://github.com/pytorch/fairseq/tree/master/examples/multilingual}.}} 

The data used for training the NMT models is available from \url{http://www.statmt.org/wmt20/quality-estimation-task.html}. We provide access to the information from the NMT model used to generate the translations: model score for the sentence and log probabilities for words, as well as the NMT systems themselves. 




\paragraph{Direct assessments.}


To collect human quality judgments, we followed the FLORES setup \cite{guzman-etal-2019-flores} inspired by the work of \citet{graham-EtAl:2013:LAW7-ID}. Specifically, the annotators were asked to rate translation quality for each sentence on a 0–100 scale, where the 0–10 range represents an incorrect translation; 11–29, a translation that contains a few correct keywords, but the overall meaning is different from the source; 30–50, a translation with major mistakes; 51–69, a translation which is understandable and conveys the overall meaning of the source but contains typos or grammatical errors; 70–90, a translation that closely preserves the semantics of the source sentence; and 91–100, a perfect translation.

Each segment was evaluated independently by three professional translators from a single language service provider. To improve annotation consistency, any evaluation in which the range of scores among the raters was above 30 points was rejected, and an additional rater was requested to replace the most diverging translation rating until convergence was achieved. To further increase the reliability of the test and development partitions of the dataset, we requested an additional set of three annotations from a different group of annotators (i.e., from another language service provider) following the same annotation protocol, thus resulting in a total of six annotations per segment.

Raw human scores were converted into z-scores, that is, standardized according to each individual annotator’s overall mean and standard deviation. The scores collected for each segment were averaged to obtain the final score. Such setting allows for the fact that annotators may genuinely disagree on some aspects of quality.

\paragraph{Human post-editing.}
For all language pairs, the translated sentences have been post-edited by human translators. For En-De and En-Zh, we used  paid editors from the Unbabel community. For all other languages, we used professional translators subcontracted by Facebook. The  human translators performing post-editing had no access to the direct assessments scores. 


\begin{table*}[!ht]
\center
\scriptsize
\begin{tabular}{p{1.2cm} p{10cm} p{2cm} }
\toprule
{\bf Type} &{\bf Text}& {\bf Scores} \\
\midrule 
{\bf Source} & He wakes up in a cage, and enjoys rubbing the rusted bars. & \\
{\bf MT} & \begin{CJK*}{UTF8}{gbsn}他在笼子里醒来, 喜欢擦生锈的酒吧. \end{CJK*} & DA = 33 \\
{\bf PE} & \begin{CJK*}{UTF8}{gbsn}他在笼子里醒来, 喜欢摩擦生锈的铁条。\end{CJK*} & HTER = 0.33 \\
\midrule
{\bf MT gloss} & He wakes up in a cage, and enjoys rubbing the rusted \textbf{pub}. & \\
{\bf PE gloss} & He wakes up in a cage, and enjoys rubbing the rusted \textbf{metal bar}. & \\
\bottomrule
\end{tabular}
\caption{Example of the discrepancy between HTER and DA annotation tasks: low DA score (low quality) but low HTER score (minimal post-editing).}
\label{tab:example-da-hter-low}

\begin{tabular}{p{1.2cm} p{10cm} p{2cm} }
\toprule
{\bf Type} &{\bf Text}& {\bf Scores} \\
\midrule 
{\bf Source} & The two battled to a standstill and eventually rendered one another comatose. & \\
{\bf MT} & \begin{CJK*}{UTF8}{gbsn}这两个人的战斗陷入停顿, 最后彼此昏迷不已. \end{CJK*} & DA = 73 \\
{\bf PE} & \begin{CJK*}{UTF8}{gbsn}两人对战陷入僵局 ， 最后双双昏倒。\end{CJK*} & HTER = 1.00 \\
\midrule
{\bf MT gloss} & The two people's battle fell into a standstill, finally both were in a coma. & \\
{\bf PE gloss} & The two people battled to a standstill and both fell into a coma. & \\
\bottomrule
\end{tabular}
\caption{Example of the discrepancy between HTER and DA annotation tasks: high DA score (high quality) but high HTER score (substantial post-editing).}
\label{tab:example-da-hter-high}
\end{table*}

\edit{Table \ref{tab:quality_stats} shows average translation quality for all language pairs based on direct assessment annotation (DA) and post-editing (HTER) for the test21 partition of the dataset. Figures \ref{fig:histograms_high_resource_test21} and \ref{fig:histograms_low_resource_test21} show the distribution of the corresponding sentence-level scores, as well as the scatter plot of DA against HTER scores.}

First, we note that the distribution of direct assessment scores is very different across language pairs. This illustrates the variety of the collected data in terms of MT output quality. For low-resource language pairs there are more sentences with low direct assessment scores, whereas in the case of high-resource language pairs the vast majority of translations received a high score. In particular, En-De has a very peaked distribution with very little variability in quality.

\edit{Second, we note that higher DA score often corresponds to lower translation edit rate in Table \ref{tab:quality_stats}. Thus, on average direct assessment and post-editing effort produce consistent results as an indication of overall translation quality per language pair.} 
However, sentence-level DA and HTER scores for the same data behave quite differently. \edit{Table \ref{tab:da_hter_correlation} shows the correlation between direct assessments and HTER scores for all the language pairs on the test21 partition of the dataset. As illustrated in Table \ref{tab:da_hter_correlation} and in the scatter plots on Figures \ref{fig:histograms_high_resource_test21} and \ref{fig:histograms_low_resource_test21} for most of the language pairs there is a weak negative correlation between the two types of quality scores.}

Direct quality assessment and post-editing give two different perspectives on MT quality. Table \ref{tab:example-da-hter-low} shows an example where direct assessment and HTER lead to a different interpretation of quality. Direct assessment score is low as the MT output contains a serious error that distorts the meaning of the sentence: ``bars'' (as in ``metal bars'') is translated as ``pub''. However the sentence is easy to post-edit as the error involves only one word to be replaced, resulting in a low HTER score. Table \ref{tab:example-da-hter-high} illustrates the opposite: MT output was assigned a high direct assessment score, but the HTER score is also high, indicating that substantial changes were introduced during post-editing. The post-edited version is more fluent, whereas the MT output is a more literal rendering of the source sentence, but the meaning is preserved and, therefore, it received a high direct assessment score.

\paragraph{Word-level labels} In the datasets containing post-edit annotation, we also obtained word-level labels for fine-grained post-editing effort estimation. Both the source and MT sides have them. 


In order to generate them, we first align source and MT outputs using SimAlign\footnote{\url{https://github.com/cisnlp/simalign}}. We follow the findings of \citet{sabet2020simalign} and use \textit{Argmax} matching for high resource languages that are close to english (En-De, En-Cs) and \textit{Itermax} for the rest of the language pairs.
We then compute the shortest edit distances between MT and post-edited texts with Tercom\footnote{\url{http://www.cs.umd.edu/~snover/tercom/}}; this effectively informs us which words were deleted, inserted or replaced. Then, any word $w_s$ in the source aligned to a word $w_m$ in MT that was kept in the post-edit receives a tag \tagok; if $w_s$ is not aligned with any other word in MT or if $w_m$ was deleted in the post-edit, it is tagged \tagbad. Thus, \tagbad{} tags in the source side indicate which words caused MT errors.

For the MT side, we tag both words and the gaps between them, indicating whether a missing additional word should have been there. Any word $w_m$ aligned to another word $w_p$ in the post-edit receives a tag \tagok; words deleted or replaced are tagged \tagbad. Any gap $g$ between words in the MT output, before the first word or after the last one receives a tag \tagok{} if no word $w_p$ is inserted in there, and \tagbad{} otherwise~\footnote{The code to reproduce the word tagging and HTER calculation for the MLQE-PE data can dbe found in \url{https://github.com/deep-spin/qe-corpus-builder}}.

Statistics for word-level tags are shown in Table~\ref{tab:word-level}. We see that most sentences in the dataset have at least one \tagbad{} tag; in the case of En-Zh, it is nearly all of them. The overall amount of \tagbad{} tags is also higher in the En-Zh data, especially in the source side.

\begin{table*}[]
    \centering
    \begin{tabular}{@{}llrrrr@{}}
    \toprule
     &  & \multicolumn{2}{c}{Source} & \multicolumn{2}{c}{Target} \\
     &  & \tagbad{} tags & Sentences & \tagbad{} tags & Sentences \\
    \midrule
    \multirow{3}{*}{En-De} & Train & 26.95\% & 92.27\% & 16.02\% & 93.60\% \\
     & Dev & 25.79\% & 91.90\% & 15.49\% & 93.40\% \\
     & Test & 25.77\% & 92.60\% & 15.53\% & 93.60\% \\
    \midrule
    \multirow{3}{*}{En-Zh} & Train & 53.59\% & 99.71\% & 30.53\% & 99.81\% \\
     & Dev & 50.92\% & 99.50\% & 28.98\% & 99.70\% \\
     & Test & 49.99\% & 99.50\% & 28.85\% & 99.70\% \\
    \bottomrule
    \end{tabular}
    \caption{Ratio of \tagbad{} tags in the word-level data for the different splits of the dataset (third and fifth columns), and ratio of sentences containing at least one such tag (fourth and sixth columns).}
    \label{tab:word-level}
\end{table*}

\section{Baseline performance}

We report the performance of baseline systems trained on the MLQE-PE data. 
\edit{The baselines trained on HTER and DA scores both follow the predictor-estimator architecture \cite{Kim2017} and are implemented using the OpenKiwi framework \cite{kepler-etal-2019-openkiwi}. The hyper-parameters used to train the baseline models are provided in Table \ref{tab:baselines-hyper-params}.

For the predictor (feature extraction) part, we use pre-trained, multilingual XLM-RoBERTa encoders \cite{conneau-etal-2020-unsupervised}. For both baselines the huggingface implementation of the XLM-RoBERTa \texttt{base} model is used~\footnote{\url{https://huggingface.co/transformers/pretrained_models.html}}. 
The \texttt{xlm-roberta-base} encoder is first fine-tuned on the concatenated source and target sentences from the train and development partitions of all language pairs (see Table \ref{tab:data_statistics}). The fine-tuning uses a masked language modeling (MLM) loss.\footnote{We use a script based on: \url{https://github.com/huggingface/transformers/blob/master/examples/legacy/run_language_modeling.py} with \texttt{per\_machine\_train\_batch\_size} set to $16$ and \texttt{block\_size} set to $512$.} The fine-tuned model is then used to jointly encode the source and target sentences, with target first. The predictor features are generated using average pooling over the target embeddings and forwarded to the estimator module which corresponds to a feed-forward layer. The combined model parameters ($\sim 281 M$ parameters) are trained on the combined training data for the DA and HTER tasks respectively ($7000$ sentence pairs for each language pair). The available combined development data ($1000$ sentence pairs for each language pair) was used to perform early stopping. Note that the configurations follow the configuration file format of OpenKiwi and any additional configurations not mentioned in Table \ref{tab:baselines-hyper-params} are identical to the default ones shown in the github configuration file.\footnote{\url{https://github.com/Unbabel/OpenKiwi/blob/master/config/xlmroberta.yaml}}}

\begin{table}[H]
    \centering
\begin{tabular}{lll}
\toprule
\textbf{Module} & \textbf{Parameter} & \textbf{Value} \\
\midrule
System & batch\_size                   & 2    \\  \hline
Encoder & hidden\_size                   & 768    \\  \hline
\multirow{2}{*}{Decoder} & dropout                        & 0.1    \\ 
& hidden\_size                   & 768    \\
Trainer & early\_stop\_patience          & 10      \\ 
\bottomrule
\end{tabular}
    \caption{Hyper-parameters for the baseline models.}
    \label{tab:baselines-hyper-params}
\end{table}


Tables~\ref{tab:baselines2021} and \ref{tab:baselines-word-2021} present the performance of our baseline systems for each label and language pair, for sentence- and word-level predictions respectively.
\begin{table}[H]
    \centering
    \begin{tabular}{lccc}
        \toprule
        Languages & Pearson $r$ & MAE & RMSE  \\
        \midrule
        \multicolumn{4}{c}{\bf Direct Assessment} \\
         En-De &0.403 &	0.629 &	0.433 \\
         En-Zh &0.525 &	0.683 &	0.534 \\
         Ru-En &0.677 &	0.702 &	0.492 \\
         Ro-En &0.818 &	0.556 &	0.408 \\
         Et-En &0.660 &	0.700 &	0.543 \\
         Ne-En &0.738 &	0.657 &	0.524 \\
         Si-En &0.513 &	0.797 & 0.626	 \\
         \midrule
         En-Cs & 0.352	& 0.845	& 0.686 \\
         En-Ja & 0.230	& 0.816	& 0.617 \\
         Km-En & 0.562	& 0.788	& 0.614 \\
         Ps-En & 0.476	& 0.852	& 0.711 \\
         \midrule
         \textbf{AVG} & 0.541 &	0.729 & 0.562 \\
        \midrule
        \multicolumn{4}{c}{\bf HTER} \\
         En-De &0.529 &	0.183 &	0.129 \\
         En-Zh &0.282 &	0.287 &	0.246 \\
         Ru-En &0.448 &	0.255 &	0.188 \\
         Ro-En &0.862 &	0.144 &	0.111 \\
         Et-En &0.714 &	0.195 &	0.149	 \\
         Ne-En &0.626 &	0.205 &	0.160 \\
         Si-En &0.607 &	0.204 &	0.159	 \\
         \midrule
         En-Cs & 0.306 & 0.262 & 0.206 \\
         En-Ja & 0.098 & 0.279 & 0.232 \\
         Km-En & 0.576 & 0.241 & 0.196 \\
         Ps-En & 0.503 & 0.333 & 0.290 \\
         \midrule
         \textbf{AVG} & 0.502 &	0.235 &	0.188 \\
        \bottomrule
    \end{tabular}
    \caption{Performance at \textbf{sentence-level} of Predictor-Estimator baseline models for each label and language pair of the \acro dataset.}
    \label{tab:baselines2021}
\end{table}
\begin{table*}[ht]
  \centering
  \begin{tabular}{lcccc|cccc}
  \toprule
    \multicolumn{1}{c}{} &\multicolumn{4}{c}{\bf Words in MT} &\multicolumn{4}{c}{\bf Words in SRC}\\
    Languages &MCC &F$_{1}$-BAD &F$_{1}$-OK &F$_{1}$-Multi   &MCC &F$_{1}$-BAD &F$_{1}$-OK &F$_{1}$-Multi \\ 
    \midrule
    En-De &0.370 & 0.455 & 0.911 & 0.415 & 0.322 & 0.393 & 0.924 & 0.363\\
    En-Zh &0.247 & 0.426 & 	0.723 &	0.308 & 0.241 &	0.394 &	0.751 &	0.295	\\
    Ru-En & 0.256 &	0.360 &	0.889 &	0.319 & 0.251 &	0.326 &	0.893 &	0.292\\
    Ro-En & 0.536 &	0.642 &	0.862 &	0.553 & 0.511 &	0.618 &	0.871 &	0.539\\
    Et-En & 0.461 &	0.589 &	0.869 &	0.512 & 0.405 &	0.522 &	0.879 &	0.459\\
    Ne-En & 0.440 &	0.828 &	0.583 &	0.483 & 0.390 &	0.768 &	0.570 &	0.438\\
    Si-En & 0.425 &	0.793 &	0.574 &	0.456 & 0.335 &	0.698 &	0.544 &	0.379\\
    \midrule
    En-Cs & 0.273 &	0.454 &	0.819 &	0.372 & 0.224 &	0.362 &	0.862 &	0.312\\
    En-Ja & 0.131 &	0.437 &	0.497 &	0.217 & 0.175 &	0.393 &	0.693 &	0.272\\
    Km-En & 0.351 &	0.766 &	0.534 &	0.409 & 0.279 &	0.644 &	0.552 &	0.355\\
    Ps-En & 0.313 &	0.674 &	0.631 &	0.425 & 0.249 &	0.501 &	0.720 &	0.361\\
    \midrule
    \textbf{AVG} & 0.346	& 0.579	& 0.717	& 0.402 & 0.307 & 0.511	& 0.751	& 0.370\\
    \bottomrule
\end{tabular}
  \caption{Performance at \textbf{word-level} of Predictor-Estimator baseline models for each label and language pair of the \acro dataset.}
	\label{tab:baselines-word-2021}
\end{table*}

\section{Conclusions}

We introduced \acro, a new dataset that was mainly created to be used for the tasks of quality estimation (sentence and word-level prediction) and automatic post-editing. It contains data in seven language pairs, direct assessment and post-editing-based sentence-level labels, as well as binary good/bad word-level labels. In addition, a subset of the data contains independently created reference translations, which can be used, for example, for machine translation evaluation. The dataset is freely available and was already used for the WMT2020 and WMT2021 shared tasks on Quality Estimation and Automatic Post-Editing.

We hope that this data will foster further work on these and other tasks, such as uncertainty estimation and model calibration. We also hope it will sparkle interest from researchers who may want to contribute related resources, i.e., more data, different languages, etc.

\section*{Acknowledgments}

Marina Fomicheva, Fr\'{e}d\'{e}ric Blain and Lucia Specia were supported by funding from the Bergamot project (EU H2020 Grant No. 825303). 
Andr\'{e} Martins and Erick Fonseca were funded by the P2020 programs Unbabel4EU (contract 042671) and MAIA (contract 045909), by  the  European  Research  Council  (ERC  StG  DeepSPIN  758969), and  by  the  Fundação  para  a  Ciência  e  Tecnologia through contract UIDB/50008/2020. We would like to thank Marina Sánchez-Torrón and Camila Pohlmann for monitoring the post-editing process. We also thank Mark Fishel from the University of Tartu for providing the Estonian reference translations.  

\clearpage
\bibliographystyle{acl_natbib}
\bibliography{references}

\begin{thebibliography}{22}
\expandafter\ifx\csname natexlab\endcsname\relax\def\natexlab#1{#1}\fi

\bibitem[{Blatz et~al.(2004)Blatz, Fitzgerald, Foster, Gandrabur, Goutte,
  Kulesza, Sanchis, and Ueffing}]{Blatz2004}
John Blatz, Erin Fitzgerald, George Foster, Simona Gandrabur, Cyril Goutte,
  Alex Kulesza, Alberto Sanchis, and Nicola Ueffing. 2004.
\newblock \href {https://www.aclweb.org/anthology/C04-1046.pdf} {{Confidence
  Estimation for Machine Translation}}.
\newblock In \emph{Proc. of the International Conference on Computational
  Linguistics}, page 315.

\bibitem[{Bojar et~al.(2014)Bojar, Buck, Federmann, Haddow, Koehn, Leveling,
  Monz, Pecina, Post, Saint-Amand, Soricut, Specia, and
  Tamchyna}]{bojar-EtAl:2014:W14-33}
Ondrej Bojar, Christian Buck, Christian Federmann, Barry Haddow, Philipp Koehn,
  Johannes Leveling, Christof Monz, Pavel Pecina, Matt Post, Herve Saint-Amand,
  Radu Soricut, Lucia Specia, and Ale\v{s} Tamchyna. 2014.
\newblock \href {http://www.aclweb.org/anthology/W/W14/W14-3302} {Findings of
  the 2014 workshop on statistical machine translation}.
\newblock In \emph{Proceedings of the Ninth Workshop on Statistical Machine
  Translation}, pages 12--58, Baltimore, Maryland, USA. Association for
  Computational Linguistics.

\bibitem[{Chatterjee et~al.(2019)Chatterjee, Federmann, Negri, and
  Turchi}]{chatterjee2019findings}
Rajen Chatterjee, Christian Federmann, Matteo Negri, and Marco Turchi. 2019.
\newblock \href {https://www.aclweb.org/anthology/W19-5402.pdf} {Findings of
  the wmt 2019 shared task on automatic post-editing}.
\newblock In \emph{Proceedings of the Fourth Conference on Machine Translation
  (Volume 3: Shared Task Papers, Day 2)}, pages 11--28.

\bibitem[{Conneau et~al.(2020)Conneau, Khandelwal, Goyal, Chaudhary, Wenzek,
  Guzm{\'a}n, Grave, Ott, Zettlemoyer, and
  Stoyanov}]{conneau-etal-2020-unsupervised}
Alexis Conneau, Kartikay Khandelwal, Naman Goyal, Vishrav Chaudhary, Guillaume
  Wenzek, Francisco Guzm{\'a}n, Edouard Grave, Myle Ott, Luke Zettlemoyer, and
  Veselin Stoyanov. 2020.
\newblock \href {https://doi.org/10.18653/v1/2020.acl-main.747} {Unsupervised
  cross-lingual representation learning at scale}.
\newblock In \emph{Proceedings of the 58th Annual Meeting of the Association
  for Computational Linguistics}, pages 8440--8451, Online. Association for
  Computational Linguistics.

\bibitem[{Espl{\`a} et~al.(2019)Espl{\`a}, Forcada, Ram{\'\i}rez-S{\'a}nchez,
  and Hoang}]{espla-etal-2019-paracrawl}
Miquel Espl{\`a}, Mikel Forcada, Gema Ram{\'\i}rez-S{\'a}nchez, and Hieu Hoang.
  2019.
\newblock \href {https://www.aclweb.org/anthology/W19-6721} {{P}ara{C}rawl:
  Web-scale parallel corpora for the languages of the {EU}}.
\newblock In \emph{Proceedings of Machine Translation Summit XVII Volume 2:
  Translator, Project and User Tracks}, pages 118--119, Dublin, Ireland.
  European Association for Machine Translation.

\bibitem[{Fomicheva et~al.(2020)Fomicheva, Specia, and
  Guzm{\'a}n}]{fomicheva2020multi}
Marina Fomicheva, Lucia Specia, and Francisco Guzm{\'a}n. 2020.
\newblock \href {https://www.aclweb.org/anthology/2020.acl-main.113.pdf}
  {Multi-hypothesis machine translation evaluation}.
\newblock In \emph{Proceedings of the 58th Annual Meeting of the Association
  for Computational Linguistics}, pages 1218--1232.

\bibitem[{Fonseca et~al.(2019)Fonseca, Yankovskaya, Martins, Fishel, and
  Federmann}]{fonseca-etal-2019-findings}
Erick Fonseca, Lisa Yankovskaya, Andr{\'e} F.~T. Martins, Mark Fishel, and
  Christian Federmann. 2019.
\newblock \href {https://doi.org/10.18653/v1/W19-5401} {Findings of the {WMT}
  2019 shared tasks on quality estimation}.
\newblock In \emph{Proceedings of the Fourth Conference on Machine Translation
  (Volume 3: Shared Task Papers, Day 2)}, pages 1--10, Florence, Italy.
  Association for Computational Linguistics.

\bibitem[{Graham et~al.(2013)Graham, Baldwin, Moffat, and
  Zobel}]{graham-EtAl:2013:LAW7-ID}
Yvette Graham, Timothy Baldwin, Alistair Moffat, and Justin Zobel. 2013.
\newblock \href {https://www.aclweb.org/anthology/W13-2305.pdf} {Continuous
  measurement scales in human evaluation of machine translation}.
\newblock In \emph{Proceedings of the 7th Linguistic Annotation Workshop and
  Interoperability with Discourse}, pages 33--41.

\bibitem[{Guzm{\'a}n et~al.(2019)Guzm{\'a}n, Chen, Ott, Pino, Lample, Koehn,
  Chaudhary, and Ranzato}]{guzman-etal-2019-flores}
Francisco Guzm{\'a}n, Peng-Jen Chen, Myle Ott, Juan Pino, Guillaume Lample,
  Philipp Koehn, Vishrav Chaudhary, and Marc{'}Aurelio Ranzato. 2019.
\newblock \href {https://doi.org/10.18653/v1/D19-1632} {The {FLORES} evaluation
  datasets for low-resource machine translation: {N}epali{--}{E}nglish and
  {S}inhala{--}{E}nglish}.
\newblock In \emph{Proceedings of the 2019 Conference on Empirical Methods in
  Natural Language Processing and the 9th International Joint Conference on
  Natural Language Processing (EMNLP-IJCNLP)}, pages 6097--6110, Hong Kong,
  China. Association for Computational Linguistics.

\bibitem[{Kepler et~al.(2019)Kepler, Tr{\'e}nous, Treviso, Vera, and
  Martins}]{kepler-etal-2019-openkiwi}
Fabio Kepler, Jonay Tr{\'e}nous, Marcos Treviso, Miguel Vera, and Andr{\'e}
  F.~T. Martins. 2019.
\newblock \href {https://doi.org/10.18653/v1/P19-3020} {{O}pen{K}iwi: An open
  source framework for quality estimation}.
\newblock In \emph{Proceedings of the 57th Annual Meeting of the Association
  for Computational Linguistics: System Demonstrations}, pages 117--122,
  Florence, Italy. Association for Computational Linguistics.

\bibitem[{Kim et~al.(2017)Kim, Lee, and Na}]{Kim2017}
Hyun Kim, Jong-Hyeok Lee, and Seung-Hoon Na. 2017.
\newblock \href {https://www.aclweb.org/anthology/W17-4763.pdf}
  {{Predictor-Estimator using Multilevel Task Learning with Stack Propagation
  for Neural Quality Estimation}}.
\newblock In \emph{Conference on Machine Translation (WMT)}.

\bibitem[{Koehn(2005)}]{koehn2005europarl}
Philipp Koehn. 2005.
\newblock \href
  {https://citeseerx.ist.psu.edu/viewdoc/download?doi=10.1.1.459.5497&rep=rep1&type=pdf}
  {{Europarl: A parallel corpus for statistical machine translation}}.
\newblock In \emph{MT summit}, volume~5, pages 79--86.

\bibitem[{Ng et~al.(2019)Ng, Yee, Baevski, Ott, Auli, and
  Edunov}]{fair-wmt2019}
Nathan Ng, Kyra Yee, Alexei Baevski, Myle Ott, Michael Auli, and Sergey Edunov.
  2019.
\newblock \href {https://arxiv.org/pdf/1907.06616.pdf} {Facebook fair wmt19
  news translation task submission}.
\newblock In \emph{Proc. of WMT}, pages 1--4.

\bibitem[{Ott et~al.(2019)Ott, Edunov, Baevski, Fan, Gross, Ng, Grangier, and
  Auli}]{ott2019fairseq}
Myle Ott, Sergey Edunov, Alexei Baevski, Angela Fan, Sam Gross, Nathan Ng,
  David Grangier, and Michael Auli. 2019.
\newblock \href {https://arxiv.org/pdf/1904.01038.pdf} {fairseq: A fast,
  extensible toolkit for sequence modeling}.
\newblock In \emph{Proceedings of the 2019 Conference of the North American
  Chapter of the Association for Computational Linguistics (Demonstrations)},
  pages 48--53.

\bibitem[{Ott et~al.(2018)Ott, Edunov, Grangier, and Auli}]{ott2018scaling}
Myle Ott, Sergey Edunov, David Grangier, and Michael Auli. 2018.
\newblock \href {https://arxiv.org/pdf/1806.00187.pdf} {Scaling neural machine
  translation}.
\newblock In \emph{Proceedings of the Third Conference on Machine Translation:
  Research Papers}, pages 1--9.

\bibitem[{Sabet et~al.(2020)Sabet, Dufter, Yvon, and
  Sch{\"u}tze}]{sabet2020simalign}
Masoud~Jalili Sabet, Philipp Dufter, Fran{\c{c}}ois Yvon, and Hinrich
  Sch{\"u}tze. 2020.
\newblock Simalign: High quality word alignments without parallel training data
  using static and contextualized embeddings.
\newblock In \emph{Findings of the Association for Computational Linguistics:
  EMNLP 2020}.

\bibitem[{Schwenk et~al.(2019)Schwenk, Chaudhary, Sun, Gong, and
  Guzm{\'a}n}]{schwenk2019wikimatrix}
Holger Schwenk, Vishrav Chaudhary, Shuo Sun, Hongyu Gong, and Francisco
  Guzm{\'a}n. 2019.
\newblock \href {https://arxiv.org/pdf/1907.05791.pdf} {{WikiMatrix: Mining
  135M Parallel Sentences in 1620 Language Pairs from Wikipedia}}.
\newblock \emph{arXiv preprint arXiv:1907.05791}.

\bibitem[{Specia et~al.(2020)Specia, Blain, Fomicheva, Fonseca, Chaudhary,
  Guzm{\'a}n, and Martins}]{specia-etal-2020-findings-wmt}
Lucia Specia, Fr{\'e}d{\'e}ric Blain, Marina Fomicheva, Erick Fonseca, Vishrav
  Chaudhary, Francisco Guzm{\'a}n, and Andr{\'e} F.~T. Martins. 2020.
\newblock \href {https://aclanthology.org/2020.wmt-1.79} {Findings of the {WMT}
  2020 shared task on quality estimation}.
\newblock In \emph{Proceedings of the Fifth Conference on Machine Translation},
  pages 743--764, Online. Association for Computational Linguistics.

\bibitem[{Specia et~al.(2018{\natexlab{a}})Specia, Blain, Logacheva, Astudillo,
  and Martins}]{specia2018findings}
Lucia Specia, Fr{\'e}d{\'e}ric Blain, Varvara Logacheva, Ram{\'o}n Astudillo,
  and Andr{\'e}~FT Martins. 2018{\natexlab{a}}.
\newblock \href {https://www.aclweb.org/anthology/W18-6451.pdf} {Findings of
  the wmt 2018 shared task on quality estimation}.
\newblock In \emph{Proceedings of the Third Conference on Machine Translation:
  Shared Task Papers}, pages 689--709.

\bibitem[{Specia et~al.(2018{\natexlab{b}})Specia, Scarton, and
  Paetzold}]{Specia2018}
Lucia Specia, Carolina Scarton, and Gustavo~Henrique Paetzold.
  2018{\natexlab{b}}.
\newblock {Quality Estimation for Machine Translation}.
\newblock \emph{Synthesis Lectures on Human Language Technologies},
  11(1):1--162.

\bibitem[{Tang et~al.(2020)Tang, Tran, Li, Chen, Goyal, Chaudhary, Gu, and
  Fan}]{Tang2020MultilingualTW}
Y.~Tang, C.~Tran, Xian Li, P.~Chen, Naman Goyal, Vishrav Chaudhary, Jiatao Gu,
  and Angela Fan. 2020.
\newblock Multilingual translation with extensible multilingual pretraining and
  finetuning.
\newblock \emph{ArXiv}, abs/2008.00401.

\bibitem[{Vaswani et~al.(2017)Vaswani, Shazeer, Parmar, Uszkoreit, Jones,
  Gomez, Kaiser, and Polosukhin}]{vaswani2017attention}
Ashish Vaswani, Noam Shazeer, Niki Parmar, Jakob Uszkoreit, Llion Jones,
  Aidan~N Gomez, {\L}ukasz Kaiser, and Illia Polosukhin. 2017.
\newblock \href {https://papers.nips.cc/paper/7181-attention-is-all-you-need}
  {Attention is all you need}.
\newblock In \emph{Advances in neural information processing systems}, pages
  5998--6008.

\end{thebibliography}
\end{document}